\documentclass[runningheads,a4paper]{llncs}

\usepackage{times}

\usepackage{balance}
\usepackage{graphicx}
\usepackage{amsmath} 		
\usepackage{amssymb}  	
\usepackage{mathrsfs}
\usepackage[usenames,dvipsnames]{color}
\usepackage[latin1]{inputenc}
\usepackage{intcalc}
\usepackage{caption}
\usepackage{float}

\newcommand {\bmilista}                   
{\begin{list} {$\bullet$} {
 \setlength{\rightmargin}{0em}
 \setlength{\leftmargin}{2em}
 \setlength{\topsep}{1ex}
 \setlength{\parsep}{0ex}
 \setlength{\parskip}{0ex}
 \setlength{\itemsep}{0ex} 
 \setlength{\listparindent}{0em}
 \setlength{\labelsep}{0.5em}
 \setlength{\labelwidth}{1em}
}}
\newcommand {\emilista} {\end{list}}

\newcommand {\Cs}{\hbox{{$\cal C$}}}                
\newcommand {\Cfree}{\hbox{{$\cal C_{\textrm{free}}$}}}                
\newcommand {\Cobstacle}{\hbox{{$\cal C_{\textrm{obs}}$}}}                

\usepackage{url}

\begin{document}

\mainmatter  

\title{Physics-based Motion Planning: Evaluation Criteria and Benchmarking}

\titlerunning{Physics-based Motion Planning: Evaluation Criteria and Benchmarking}

%
%
\author{Muhayyuddin, Aliakbar Akbari and Jan~Rosell%
\thanks{This work was partially supported by the Spanish Government through the projects \mbox{DPI2011-22471}, \mbox{DPI2013-40882-P} and \mbox{DPI2014-57757-R}.
Muhayyuddin is supported by the Generalitat de Catalunya through the grant FI-DGR 2014.}
}
\authorrunning{Muhayyuddin, A. Akbari and J.~Rosell}

\institute{Institute of Industrial and Control Engineering,\\
 Universitat Polit\`ecnica de Catalunya, Barcelona, Spain,\\
\email{\{muhayyuddin.gillani,aliakbar.akbari,jan.rosell\}@upc.edu} 
    }

%
%

\maketitle

\begin{abstract}
Motion planning has evolved from coping with simply geometric problems to physics-based ones that incorporate the kinodynamic and the physical constraints imposed
by the robot and the physical world. Therefore, the criteria for evaluating physics-based motion planners goes beyond the computational complexity 
(e.g. in terms of planning time) usually used as a measure for evaluating geometrical planners, in order to consider also the quality of the solution in terms of dynamical parameters.
This study proposes an evaluation criteria and analyzes the performance of several kinodynamic planners, which are at the core of physics-based motion planning, using different scenarios with fixed and manipulatable objects.
RRT, EST, KPIECE and SyCLoP are used for the benchmarking. The results show that KPIECE computes the time-optimal solution with heighest success rate, whereas, 
SyCLoP compute the most power-optimal solution among the planners used.

\end{abstract}

\section{Introduction} \label{s-Introduction}
Robotic manipulation requires precise motion planning and control to execute the tasks, either for industrial robots, mobile manipulators, or humanoid robots.
It is necessary to determine the way of safely navigating the robot from the start to the goal state by satisfying the kinodynamic (geometric and differential)
constraints, as well as to incorporate the physics-based constraints imposed by possible contacts and by the dynamic properties of the world such as gravity and friction~\cite{tsianos2007,ladd2005}.
These issues  significantly increase the computational complexity because certain collision-free geometric paths may not be feasible in the presence of these constraints.

Physics-based motion planning has emerged, therefore, as a new class of planning algorithms that considers the physics-based constraints along with the
kinodynamical constraints, i.e. it is an extension to the kinodynamic motion planning~\cite{zickler2010} that also involves the purposeful manipulation
 of the objects by considering  the dynamical interaction between rigid bodies. This interaction is simulated based on the principal
of basic Newtonian physics and the results of simulation are used for planning. The performance of a physics-based planner largely depends on the choice
of the kinodynamic motion planner, that is implicitly used for sampling the states and the construction of the solution path. The  state propagation is performed using
dynamic engines, like ODE~\cite{OpenDE2007}, that incorporates the kinodynamical and physics-based constraints.

Motion planning in its simplest form (i.e. as a geometric problem) is PSPACE-complete~\cite{reif1979}. The incorporation
of kinodynamic constraints and the physics-based properties make it even more complex and computationally intensive, and for complex systems
even the decidability of the physics-based planning problem is questionable~\cite{cheng2007}.
Therefore, to make the physics-based planning computationally tractable, it is crucial to use the most appropriate and computationally efficient
kinodynamic planner. In previously proposed physics-based planning approaches, different kinodynamic motion planners and physics engines have been used.

A few studies provided comparative analysis of the performance of some kinodynamical motion planners within   different  physics-based planning frameworks.
For instance, the physics-based planning algorithm proposed in~\cite{zickler2009}, that used nondeterministic tactics and skills to reduce the search space of
physics-based planning, was evaluated (in term of planning time and tree length) using two different kinodynamic motion planners,
Behavioral Kinodynamic Rapidly-exploring Random Trees (BK-RRT~\cite{zickler20}) and  Balanced Growth Trees (BGT~\cite{zickler2009}).
The physics engine PhysX~\cite{physx} was used as
state propagator.
Another physics-based planning approach~\cite{plaku2012} integrated the sampling-based motion planing
with the discrete search using the workspace decomposition in order to map the planning problem onto a graph searching problem. This work
evaluated the performance (in term of planning time) using  RRT, Synergistic Combination of Layers of Planning (SyCLoP~\cite{Plaku2010}) and a modified version of the SyCLoP
as kinodynamic motion planners. The propagation step was performed using the Bullet~\cite{bullet2013} physics engine.
A third approach  proposed a physics-based motion planning framework that used manipulation knowledge coded as an ontology~\cite{muhayyuddin2015}.
This approach performed a reasoning process over the knowledge to improve the computational efficiency and has shown a significant
improvement in performance (in term of planning time and generated trajectory), as compared to the simple physics-based planning.
Two kinodynamic motion planners were used, Kinodynamic Planning by Interior-Exterior Cell Exploration (KPIECE~\cite{sucan2012}) and RRT. The Open Dynamics Engine (ODE) was used
as state propagator.

All the above stated studies basically measured the time complexity of different kinodynamic motion planners. Since physics-based planning
simultaneously evaluates the kinodynamical and the physics-based constraints, the evaluation based on just planning time may not be sufficient.
New evaluation criteria is required because a number of other dynamical parameters (such as power consumed, action, smoothness) may significantly
influence the planning decisions, like in the task planning
approaches proposed in~\cite{Ali2015,RobotAli2015}  that use a physics-based reasoning process to determine the feasibility of a plan by evaluating the cost in terms of
power consumed and the action.
With this in mind, the present study proposes a new benchmarking
criteria for the physics-based planning that incorporates the dynamical properties of the system (to determine the
quality of the solution) as well as the computational complexity. It is used to compare different kinodynamic motion planners
(RRT, EST, KPIECE and SyCLoP)
within the physics-based planning framework presented in~\cite{muhayyuddin2015}  based on a reasoning process over ontological manipulation knowledge.

\section{Kinodynamic Motion Planning}\label{s-MotionPlanning}
Motion planning problems deal with computing collision-free trajectories from a given start to a goal state in the configuration space
(\Cs), the set of all possible configurations of the robot~\cite{Perez1983}.
The geometrically accessible region of \Cs\ is called \Cfree\ and the obstacle region is known as \Cobstacle.
The sampling-based algorithms such as Probabilistic RoadMaps~\cite{Kavraki1996} and the Rapidly-exploring Random Trees~\cite{lavalle2000} have shown  significant
performance when planning in high-dimensional configuration spaces. These algorithms connect collision-free configurations
 with either a graph or a tree to capture the connectivity of \Cfree\ and find a path along these data structures to connect the initial and the goal configurations.

Kinodynamic motion planning refers to the problems in which the motion of the robot must simultaneously satisfy the kinematic constraints (such as joint limits
and obstacle avoidance) as well as some dynamic constraints (such as bounds on the applied forces, velocities and accelerations~\cite{donald1993}).
Tree-based planners are best suited to take into account kynodynamic constraints~\cite{tsianos2007}, since the  dynamic equations are
used to determine the resulting motions used to grow the tree.
The general functionality of sampling-based
kinodynamic planners is to search a state space $S$ of higher dimensions that records the system's dynamics. The state of a robot for a configuration
$q \in$ \Cs\ is defined as $s=(q,\dot{q})$. To determine a solution, the planning will be performed in state space, in a similar way
as in \Cs. This section briefly reviews the existing most commonly used kinodynamic motion planners, that can be categorized into three classes:
a) RRT and EST belong to the class
of planning algorithms that
sample the states; b) KPIECE belongs to the class that samples the motions or path segments; c) SyCLoP is an hybrid planner
that splits the planning problem into a discrete and a continuous layer.
\vspace{-4mm}
\subsubsection{Rapidly Exploring Random Trees (RRT):}
It is a sampling-based kinodynamic motion planning algorithm~\cite{lavalle2001} that has the ability to efficiently explore the high dimensional configuration
spaces. The working mechanism of RRT-based algorithms is to randomly grow a tree rooted at the start state ($q_{start}\in $ \Cs), until it finds
 a sample at the goal state ($q_{goal} \in $ \Cs). The growth of the tree is based on two steps, \textit{selection} and \textit{propagation}. In the
first step a sample is randomly selected ($q_{rand}$), and its nearest node in the tree is then searched ($q_{near}$). The second step applies, from $q_{near}$,
random controls (that satisfy the constraints) during a certain amount of time. Among the configurations reached, the one nearest to $q_{near}$ is selected as $q_{new}$ and an edge from
$q_{near}$ to $q_{new}$ is added to the tree. Using this procedure, all the paths on the tree will be feasible, i.e. by construction they satisfy all the kinodynamic
constraints.
\vspace{-4mm}
\subsubsection{Expansive-Spaces Tree planner (EST):}
This approach constructs a tree-shaped  road-map~$T$ in the \textit{state$\times$time} space~\cite{hsu1997}~\cite{hsu200}. The idea is to select a milestone
of $T$ and from there randomly apply sampled controls for a certain amount of time. If the final state is in free-space, it will be added as a
milestone in $T$. The selection of the milestone for expansion is done in a way that the resultant tree should neither be too dense nor sparse.
This kinodynamic planner works in three steps: \textit{milestone selection}, \textit{control selection} and \textit{endgame connection}.
In the first step, a milestone $m$ in $T$ is selected with probability inversely proportional to the
 number of neighboring milestones of $m$.
In the second step, controls are randomly sampled and applied from the selected milestone $m$. Since by moving under kinodynamic constraints
it may not be possible to reach exactly the goal state, the \textit{endgame connection} step is the final step that defines
a region around the goal in such a way that any milestone within this region is considered as the goal state.
\vspace{-4mm}
\subsubsection{Kinodynamic Motion Planning by Interior-Exterior Cell Exploration (KPIECE):}
This planner is particularly designed for  complex dynamical systems.
KPIECE grows a tree of motions by applying randomly sampled controls for a randomly sampled time duration from a tree node selected as follows.
The state space is projected onto a lower-dimensional space that is partitioned into cells in order to estimate the coverage.
As a result of this projection, each motion will be part of a cell, each cell being classified as an interior or exterior cell depending on whether the neighboring cells are occupied or not.
Then, the selection of the cell is performed based on the \textit{importance} parameter that is computed based on: 1) the \textit{coverage} (the cells that are less covered are preferred over
the others); 2) the \textit{selection} (the cells that have been selected less number of times are preferred); 3) the \textit{neighbors} (the cells that have less neighbors are preferred); 4) the \textit{selection time} (recently selected cells are preferred); 5) the \textit{expansion} (easily expanded cells are preferred over the cells that expand slowly).
The cell that has maximum importance will be selected.
The process continues until the tree of motion reaches the goal region.
\vspace{-4mm}
\subsubsection{Synergistic Combination of Layers of Planning (SyCLoP)}
This is a meta approach that considers motion planning as a search problem in a hybrid space (of a continuous and a discrete layer) for efficiently solving the problem
under kinodynamical constraints. The continuous layer is represented by the state space (that is explored by a sampling-based motion planner like RRT or EST) and the discrete layer is determined by the decomposition of the
workspace. The decomposition is used to compute a cost parameter called \textit{lead} that guides the motion planner towards the goal. SyCLoP works based on the following
steps: \textit{lead computation} and \textit{region selection}. The lead is computed based on the coverage and the frequency of the selection. The former is obtained by
the sampling-based motion planner (continuous layer) and the latter is computed by determining how many
times a cell has been selected from discrete space. The selection of the region will be performed based on the available free volume of the region (high free volume
regions are preferred for the exploration). The process continues until the planner finds a sample in the goal region.
SyCLoP will be recalled SyCLoP-RRT or SyCLoP-EST based on the planner used in the continuous layer.
\subsection{Ontological Physics-Based Motion Planning}
Physics-based motion planning is composed of a new class of planning algorithms that basically go a step further towards physical realism, by also taking into account possible interactions between bodies and
possible physics-based constraints (such as gravity and friction that conditions the actions and its results).
The search of collision-free trajectories is not the final aim; now collisions with some objects may be allowed, i.e. these  algorithms also consider the manipulation actions (such as push action) in order
to compute the appropriate trajectory.
 The incorporation of the dynamic interaction (for manipulation) between rigid bodies and other physics-based constraints
 increase the dimensionality of the state space and the computational complexity. In some cases, particularly for the systems with
complex dynamics, the problem may even be not tractable.

The ontological physics-based motion planning is a recently proposed approach that tries to cope with aforementioned challenges~\cite{muhayyuddin2015}. This approach takes the advantage
of Prolog-based reasoning process over the knowledge  of objects and manipulation actions (this knowledge is represented in the form of an ontology).
The reasoning process is used to improve  the computational efficiency and to make the manipulation problem computationally tractable. It applies a hybrid approach consisting of two main layers which are
 a knowledge-based reasoning layer and the motion planning layer.

The knowledge-based reasoning layer uses the manipulation ontology to derive a knowledge, called abstract knowledge, that contains information of the objects and their properties (such as their manipulatable regions, e.g. the regions from where an object can be pushed),
 and the initial and goal
states of the robot. The abstract knowledge, moreover, categorizes the objects into fixed and manipulatable objects, being the manipulatable objects further divided
into freely and constraint-oriented manipulatable ones (e.g. some objects can be pushed from any region while others may only be pushed from some given region and in some predefined directions).
 Furthermore, abstract knowledge also determines the geometrical positions of the objects to distinguish
whether the goal state is occupied or not.

The motion planning layer includes a reasoning process that infers from the abstract knowledge. The inferred knowledge
is called instantiated knowledge and is a dynamic knowledge that is updated at each instance of time. Motion planning layer employs a sampling-based
kinodynamic motion planner (like KPIECE or RRT) and a physics engine used as state propagator. After the propagation step, the new state is accepted by the
planner if all the bodies satisfy the manipulation constraints imposed by the instantiated knowledge (e.g. a car-like object can only be pushed forward or backward and therefore any state that results with a collision with its lateral sides is disallowed). In this way the growing of the tree-like data structure of the planner is more efficient since useless actions are pruned.

\section{Benchmarking Parameters}\label{s-benchmarking}
A wide variety of the kinodynamic motion  planners is available, the planning strategy of these algorithms is conceptually different
from one another, and this can significantly affect the performance of the physics-based planning. A criteria is proposed to evaluate the performance of the
kinodynamic motion planners
for the physics-based planning. It is suggested that, along with the computational complexity, it is also necessary to evaluate the dynamical parameters
that determine the quality of
the computed solution path. This is determined by estimating the power consumed by the robot while moving along the computed path, by the
total amount of action (i.e. dynamic attribute of trajectory explained later in this section) of the computed trajectory and by the smoothness of the trajectory. The computational complexity is 
computed based on the
planning time and the average success rate of each planner. A planner is said to be most appropriate if it is optimal according to the above said criteria.

The choice of physics engine (such as ODE, Bullet and PhysX) may not affect the simulation results because the
design philosophy of all of them is based on the basic physics, and the performance and accuracy may vary a little but the simulation results of all the physics engines are  almost the same.

The following performance parameters have been established to evaluate different kinodynamic motion planners within the framework of the ontological physics-based motion planner.
The trajectories given by the kinodynamic motion planners are described by a list of forces and their duration that have to be consecutively applied to move the robot (either in a collision-free way or possibly pushing some manipulatable objects):
\begin{itemize}
 \item \textbf{\textit{Action:}}
 It is a dynamical property of a physical system, defined in the form of a functional $\mathcal{A}$
 that takes a sequence of moves that define a trajectory as input, and returns a scalar number as  output:
 \begin{equation}
\label{eq:action}
  \mathcal{A}  =\sum\limits_{i}^n|\mathbf{f_i}| \Delta t_i \varepsilon_i,
\end{equation}
where $\mathbf{f}_i$, $\Delta t_i$ and $\varepsilon_i$ are, respectively, the applied control forces, their duration and the resulting covered distances.
 \item \textbf{\textit{Power consumed:}}
 The total amount of power consumed $\mathcal{P}$ by the robot to move from start to the goal state is computed as:
 \begin{equation}
\label{eq:power}
  \mathcal{P}  =  \sum\limits_{i}^n\frac{\mathbf{f_i} \mathbf{d_i}}{\Delta t_i},
\end{equation}
where $\mathbf{f}_i$, $\mathbf{d}_i$ and $\Delta t_i$ are, respectively, the applied control forces, the resultant displacement vectors and the time duration.
 \item \textbf{\textit{Smoothness:}} The smoothness $\mathcal{S}$ of a trajectory can be measured as a function of jerk~\cite{hogan1984}, the time derivative of
 acceleration:
  \begin{equation}
\label{eq:jerk}
\mathcal{J}(t) =   \frac{d\,a(t)}{dt}.
\end{equation}
 For a given trajectory $\tau$ the smoothness is defined as the sum of squared jerk along~$\tau$:
 \begin{equation}
\label{eq:smoothness}
\mathcal{S} =   \int_{t_i}^{t_f} \mathcal{J}(t)^2\,dt,
\end{equation}
where $t_i$ and $t_f$ are the initial and final time, respectively.
 \item \textbf{\textit{Planning time:}}
 It is the total time consumed by the ontological physics-based planner to compute a solution trajectory.
 \item \textbf{\textit{Success Rate:}}
It is computed based on the number of successful runs.
\end{itemize}

\section{Results and Discussion}
In this section  the results of benchmarking of kinodynamical motion planners for ontological physics-based planning is  presented. The benchmarking is performed using
\textit{The Kautham Project}~\cite{Rosell2014},  a C++ based open source platform for  motion planning that includes geometric, kinodynamic, and ontological physics-based motion planners.
It uses the planning algorithms offered by the Open Motion Planning Library (OMPL)~\cite{sucan201}, an  C++ based open source motion planning library, and ODE as state
propagator for the physics-based planning.

\subsection{Simulation Setup}
\begin{figure}[!h]
\begin{center}
\includegraphics[scale = 0.33]{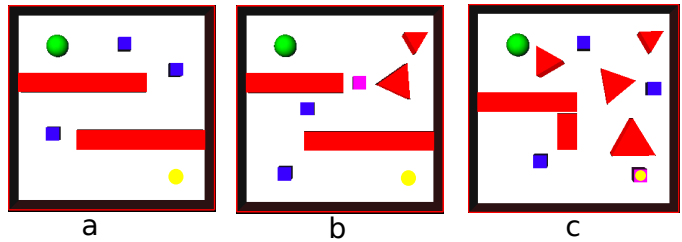}
\caption{Simulation setup  used for the benchmarking. The robot (green sphere) is depicted at the start configuration while the yellow circle represents the goal.}\label{fig:scenes}
\vspace{-6mm}
\end{center}
\end{figure}

The simulation setup consists of a robot (green sphere), free manipulatable bodies (blue cubes), constraint oriented manipulatable bodies
(purple cubes), and the fixed bodies
(triangular prisms, walls, and floor). The benchmarking is performed with the three different scenes shown in Fig. \ref{fig:scenes}, that are
differentiated based on the degree of clutter. The robot is depicted at its initial configuration, being the goal robot configuration painted as a yellow circle.
Fig. \ref{fig:scenes}-a describes the simplest scene that consists of a robot, free manipulatable bodies, and fixed bodies. The second scene, represented in
Fig.\ref{fig:scenes}-b, consists of a robot, free manipulatable bodies, fixed bodies, and a constraint-oriented body. In this scene the narrow passage is occupied by the
constraint-oriented manipulatable body (it can only be pushed vertically, along y-axis) that has to be pushed away by the robot in order to clear the path towards the goal.
It is important to note that since there is not any collision-free path available from the start to the goal state, the geometric as well as the kinodynamic planners
are not able to compute the path; only physics-based planners has the ability to compute the path by pushing the object away. The final scene is depicted in Fig.~\ref{fig:scenes}-c and
has the highest degree of clutter. The goal is occupied with a constraint-oriented manipulatable body (it can only be pushed horizontally, along \hbox{x-axis);} 
in order to reach the goal region the robot needs to free it by pushing the body away. As before, no collision-free path exists. The same planning parameters
are used for all the planners: goal bias equal to 0.05, sampling control range between -10N and 10N, and propagation step size equal to 0.07s.
\vspace{-4mm}
\subsection{Benchmarking Results}

\begin{figure}[!h]
\begin{center}
\includegraphics[scale = 0.27]{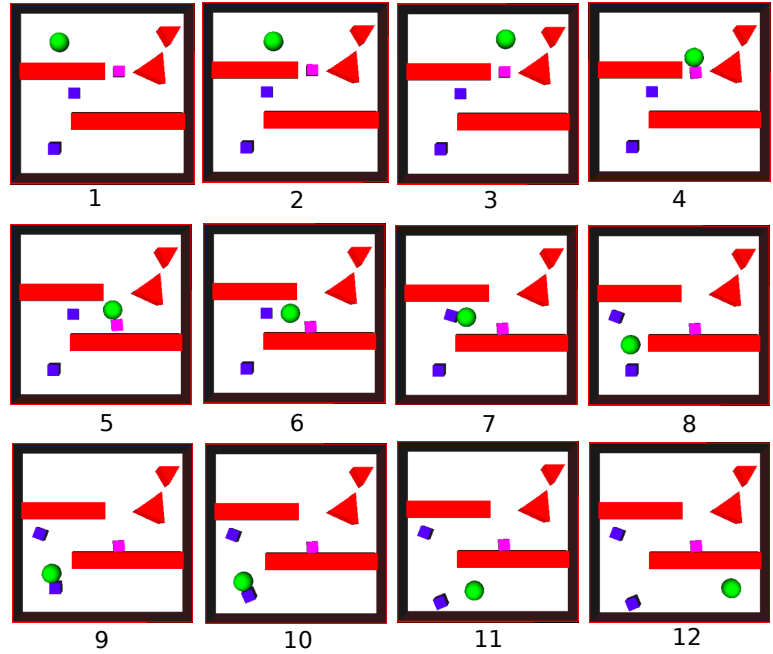}
\caption{Sequence of snapshots of the execution using ontological physics-based motion planner.}\label{fig:run}
\vspace{-4mm}
\end{center}
\end{figure}

\begin{figure}[!h]
\begin{center}
 \includegraphics[scale = 0.20]{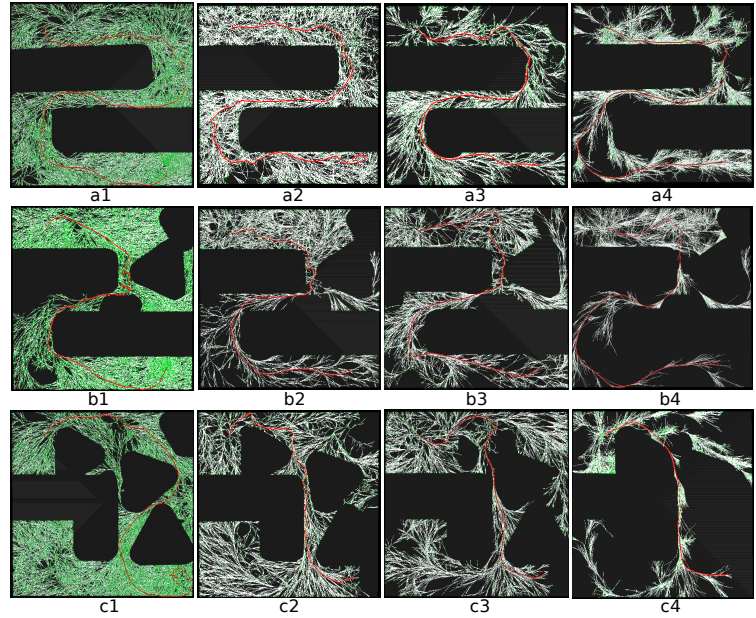}
\caption{Configuration space and solution path: each row corresponding to the scene, the columns in each row (left to right)
 represents the configuration space and solution path using
KPIECE, RRT, SyCLoP-RRT, and SyCLoP-EST respectively.}\label{fig:cspace}
\vspace{-7mm}
\end{center}
\end{figure}

\begin{figure}[!ht]
\begin{center}
\includegraphics[height=4cm,width=9cm]{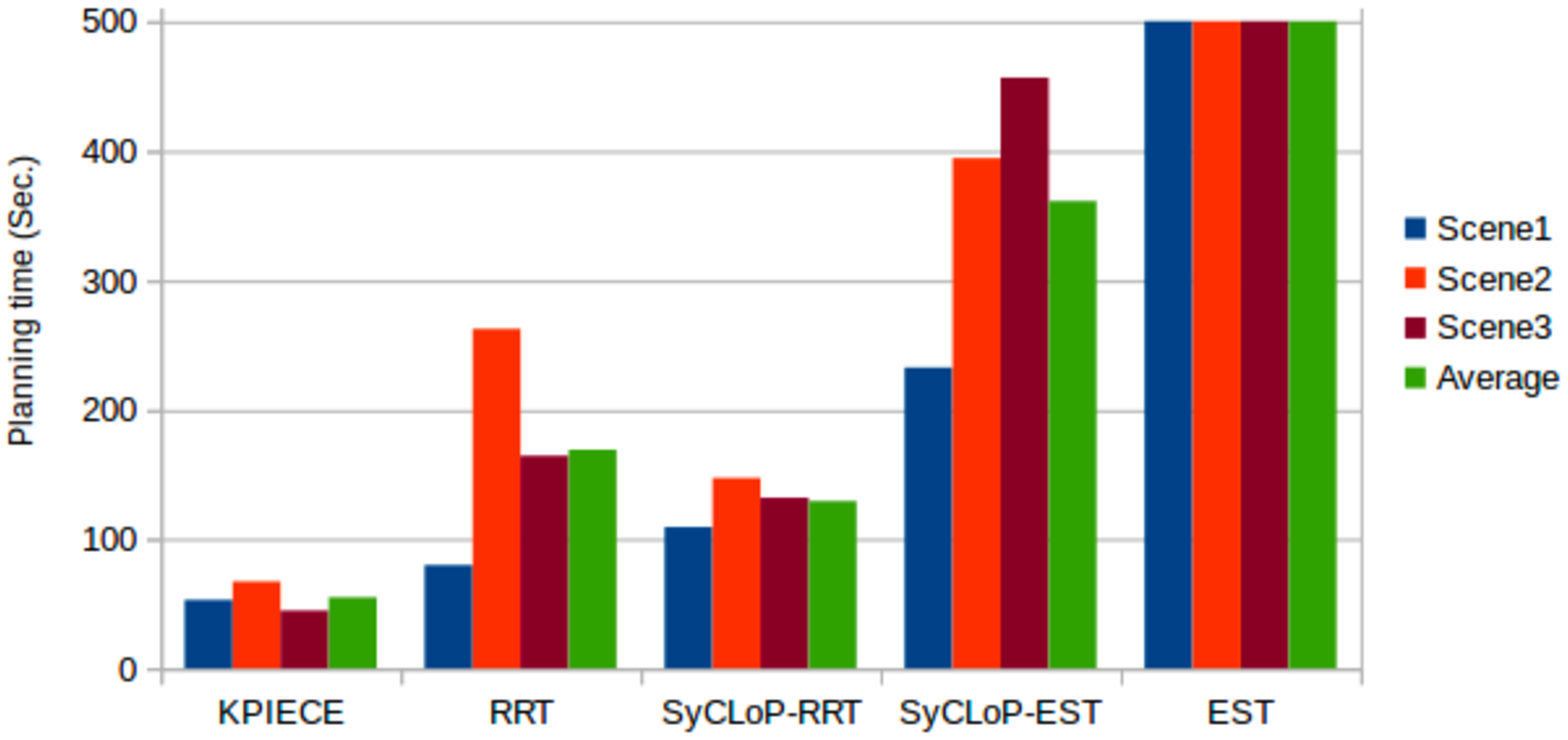}\\
\caption{Average planning time (10 runs) for each scene and the overall averaged planning time (three scenes).} \label{fig:time}
\vspace{4mm}
\includegraphics[height=4cm,width=9cm]{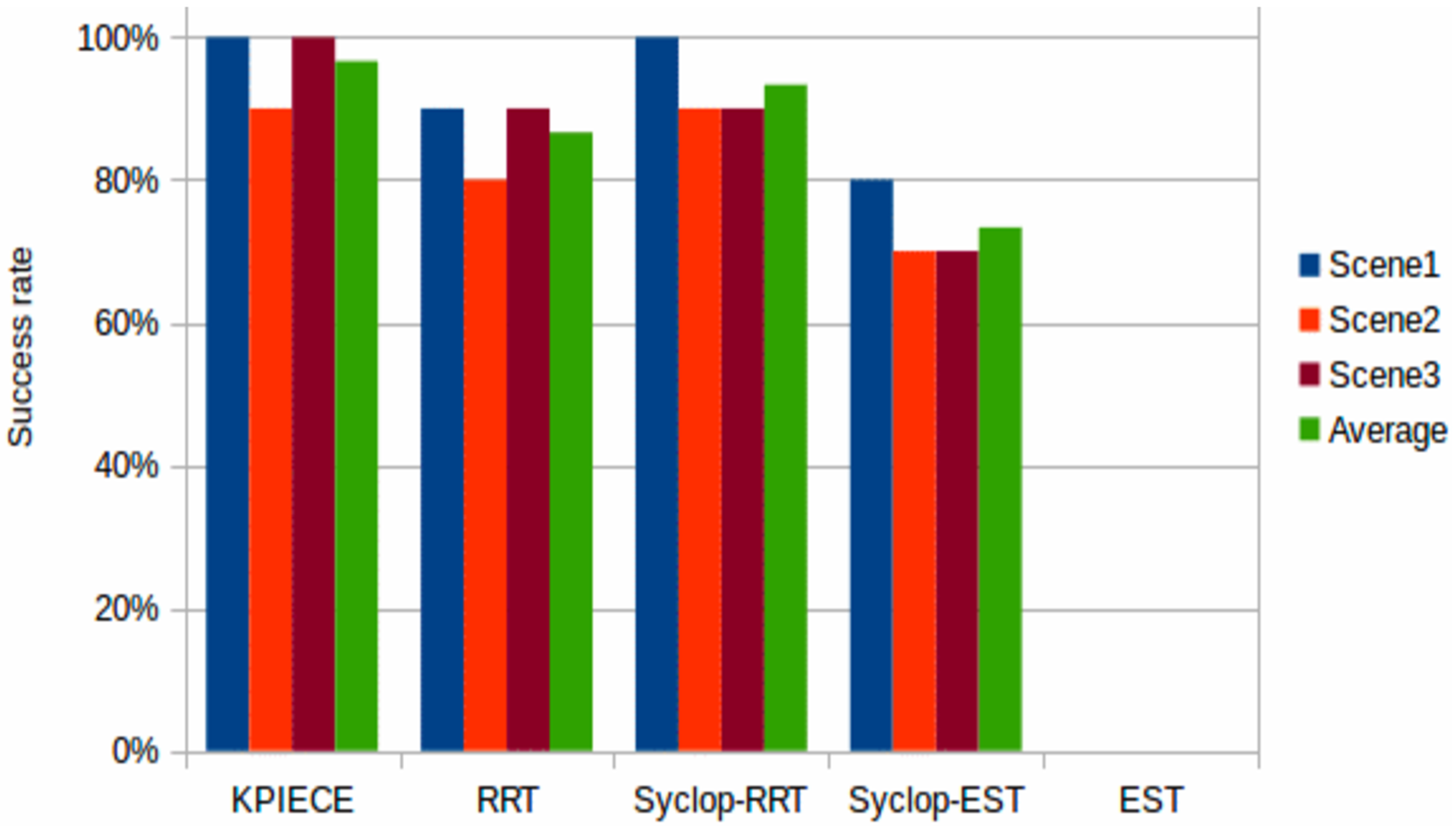}
\caption{Success rate of the planners for each scene and the overall averaged success rate (three scenes).} \label{fig:asr}
\end{center}
\end{figure}

The ontological physics-based motion planner is run 10 times for each scene and for each of the kinodynamic motion planners summarized in
Section~\ref{s-MotionPlanning}. The average values of the benchmarking parameters are presented
in the form of histograms. Fig.~\ref{fig:run} shows as sequence of snapshots of a sample execution of the ontological physics-based motion planner using KPIECE.
In order to estimate the coverage of configuration space, and the solution trajectory Fig.~\ref{fig:cspace} depicts the configuration spaces and solution paths
computed by different kinodynamic planners.
Fig.\ref{fig:time} shows the average planning time (being the maximum allowed planning time set to 500 s). All the
planners have been able to compute the solution within the maximum range of planning time except EST. Among all the planners KPIECE computed the solution most efficiently.
\begin{figure}[!ht]
\begin{center}
\includegraphics[height=4cm,width=8cm]{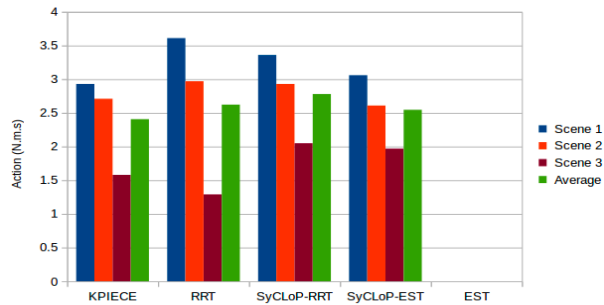}
\caption{Average amount of action for each scene and the overall averaged  amount of action (three scenes).} \label{fig:action}
\vspace{-6mm}
\end{center}
\end{figure}
\begin{figure}[!ht]
\begin{center}
\includegraphics[height=3.7cm,width=8cm]{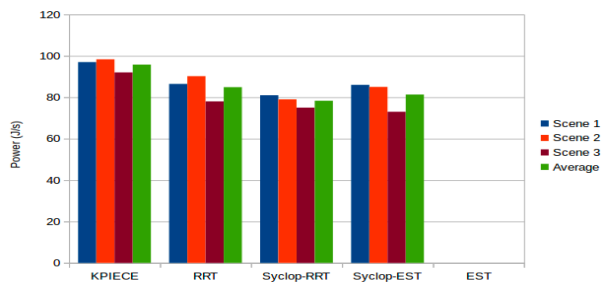}
\caption{Average power consumed by the robot while moving along the solution path and the overall average (three scenes).} \label{fig:power}
\vspace{-5mm}
\end{center}
\end{figure}
\begin{figure}[!ht]
\begin{center}
\includegraphics[height=3.7cm,width=8cm]{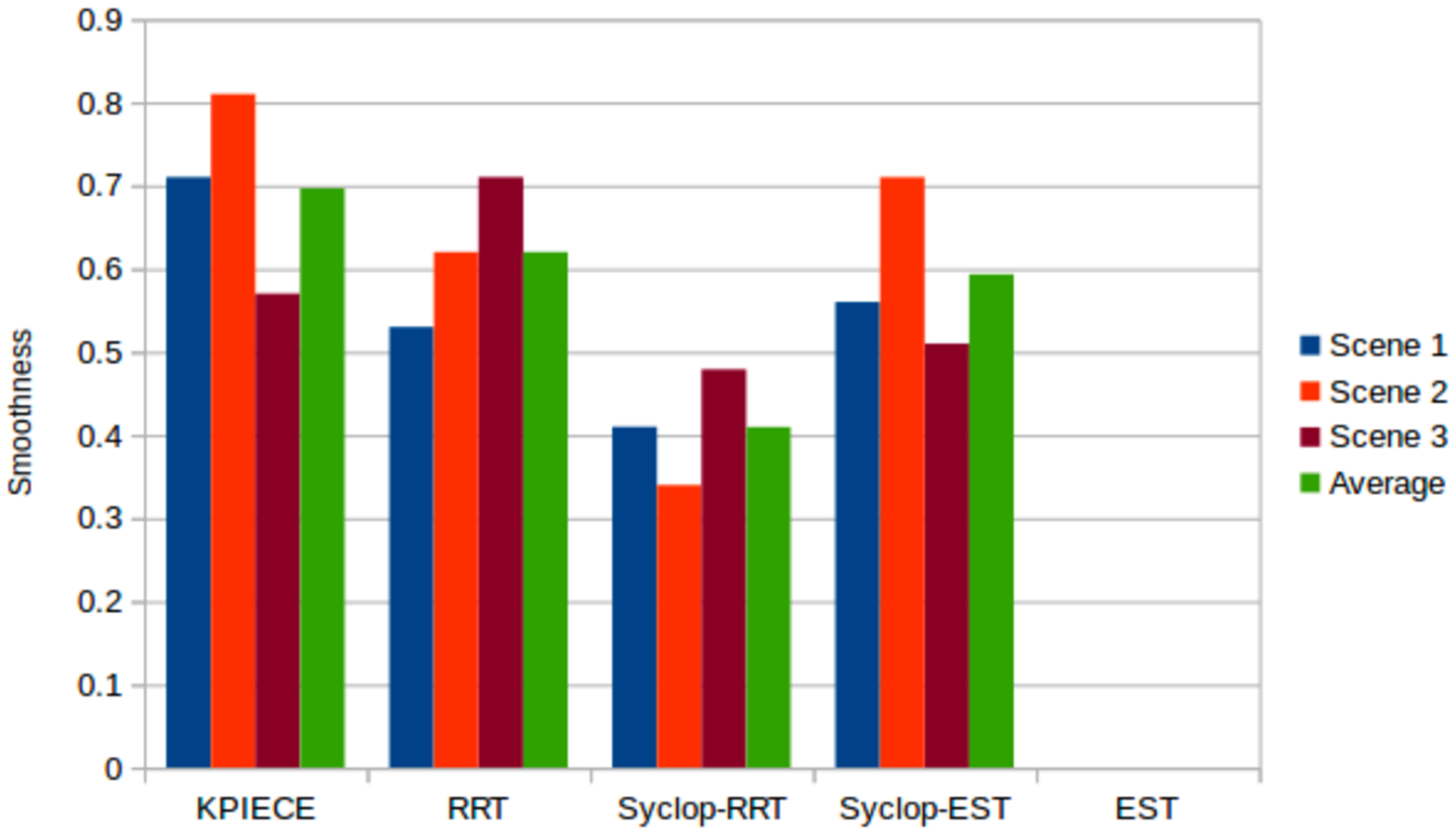}
\caption{Smoothness measure for each scene and the overall average  (three scenes). } \label{fig:smooth}
\vspace{-4mm}
\end{center}
\end{figure}
The success rate of the planners is described in Fig.~\ref{fig:asr}. It is computed for each scene individually based on the number of successful runs (i.e. how many times the planner
computes the solution within the maximum limit of time), then
the average success rate of each planner is determined by computing the average successful runs for the three scenes. Results show that the KPIECE has the highest
overall success rate. The SyCLoP-RRT also shows an impressive success rate, whereas, EST has zero success rate.

The results of the dynamical parameters, action and power, are shown in Fig.~\ref{fig:action} and Fig.~\ref{fig:power}, respectively. 
Regarding action, the solution trajectory
with minimum amount of action is considered as the most appropriate. Among the planners that computed the solution within the given time, on average, KPIECE
has the minimum amount of action and SyCLoP-RRT has the maximum one. 
Regarding power, it is desirable that the robot should consume a minimum amount of power while moving along the
solution. Our analysis shows that SyCLoP-RRT finds the power-optimal trajectory whereas KPIECE is the worst. The results for
RRT and SyCLoP-EST are almost the same.  
Regarding trajectory smoothness, Fig.~\ref{fig:smooth} shows the comparison. The SyCLoP-RRT computes the most smooth trajectory among all the planners
whereas, KPIECE show the worst results in term of smoothness. Since EST was not able to compute the solution within the given time, the action, power and smoothness
for EST are set to infinity and not shown in the histograms.

\subsection{Discussion}
We proposed a benchmarking criteria for physics-based planning. Based on the proposed criteria, the performance of physics-based planning is evaluated using
different kinodynamic motion planners. Our analysis shows that in terms of planning time, success rate and the action value, KPIECE is the most suitable planner. SyCLoP-RRT
shows significant results in terms of smoothness of the computed trajectory and power consumed by the robot moving along the computed path. The planning
time and success rate for SyCLoP-RRT is also impressive. RRT shows average results throughout the evaluation. SyCLoP-EST is good in terms of action value and power
consumption but the value of planning time is very high and it has a low success rate. The EST was not able to compute the solution within time and has zero success rate.

\section{Conclusion and Future Work}
This paper proposed an evaluation criteria for the physics-based motion planners. The proposed benchmarking criteria computes  dynamical parameters
(such as power consumed by the robot to move along the solution path) for the evaluation of the quality of the computed solution path.
Further, based on the proposed benchmarking criteria, the performance of the ontological physics-based motion planner (using different kinodynamic motion
planners) is evaluated and the computed properties of the each kinodynamic motion planner are discussed in detail. For now the evaluation criteria
was implemented on simple scenes and with the push action as the sole manipulation action; as future work the proposed benchmarking criteria will be implemented for mobile manipulators to also benchmark the grasping
and pick and place manipulation actions.

\balance
\bibliographystyle{splncs}
\bibliography{References}
\end{document}